\documentclass{bmvc2k}

\usepackage{enumitem}
\usepackage{array}
\usepackage{balance}
\usepackage{color}
\usepackage[toc,page]{appendix}
\usepackage{bbm}
\usepackage{algorithm}
\usepackage[noend]{algpseudocode} 
\usepackage{bbding}
\usepackage{booktabs}
\usepackage[normalem]{ulem}
\usepackage{setspace}
\usepackage{stfloats}
\usepackage{bm}

\usepackage{amsmath}
\usepackage{amsfonts}       
\usepackage{wrapfig}
\usepackage{sidecap}

\definecolor{color}{rgb}{0.05, 0.5, 0.15}

\newcolumntype{C}[1]{>{\centering\let\newline\\\arraybackslash\hspace{0pt}}m{#1}}

\addauthor{Amin Banitalebi-Dehkordi}{amin.banitalebi@huawei.com}{1}
\addauthor{Xinyu Kang}{xinyu.kang@alumni.ubc.ca}{2}
\addauthor{Yong Zhang}{yong.zhang3@huawei.com}{1}

\addinstitution{
 Huawei Technologies Canada Co., Ltd.
}
\addinstitution{
 University of British Columbia,\\
 Vancouver, Canada
}

\runninghead{Banitalebi-Dehkordi, Kang, and Zhang}{Model Composition}

\begin{document}

\title{Model Composition: Can Multiple Neural Networks Be Combined into a Single Network Using Only Unlabeled Data?}

\maketitle

\begin{spacing}{0.94}

\vspace{-18pt}
\begin{abstract}
\vspace{-4pt}
The diversity of deep learning applications, datasets, and neural network architectures necessitates a careful selection of the architecture and data that match best to a target application. As an attempt to mitigate this dilemma, this paper investigates the idea of combining multiple trained neural networks using unlabeled data. In addition, combining multiple models into one can speed up the inference, result in stronger, more capable models, and allows us to select efficient device-friendly target network architectures. To this end, the proposed method makes use of generation, filtering, and aggregation of reliable pseudo-labels collected from unlabeled data. Our method supports using an arbitrary number of input models with arbitrary architectures and categories. Extensive performance evaluations demonstrated that our method is very effective. For example, for the task of object detection and without using any ground-truth labels, an EfficientDet-D0 trained on Pascal-VOC and an EfficientDet-D1 trained on COCO, can be combined to a RetinaNet-ResNet50 model, with a similar mAP as the supervised training. If fine-tuned in a semi-supervised setting, the combined model achieves +18.6\%, +12.6\%, and +8.1\% mAP improvements over supervised training with 1\%, 5\%, and 10\% of labels. Code is released as supplementary \cite{supplementary}.
\end{abstract}

%

\vspace{-8pt}
\section{Introduction} \label{introduction}
\vspace{-4pt}
Deep learning has enabled achieving outstanding results on a wide range of applications in computer vision and image processing \cite{alom2019state, shrestha2019review}. However, the diversity of datasets and neural network architectures necessitates a careful selection of model architecture and training data that match best to the target application. Often times, for a same task, many models are available. These models might be trained on different datasets, or might come in different capacities, architectures, or even bit precisions. 

\begin{SCfigure}
\centering\includegraphics[width=0.58\paperwidth]{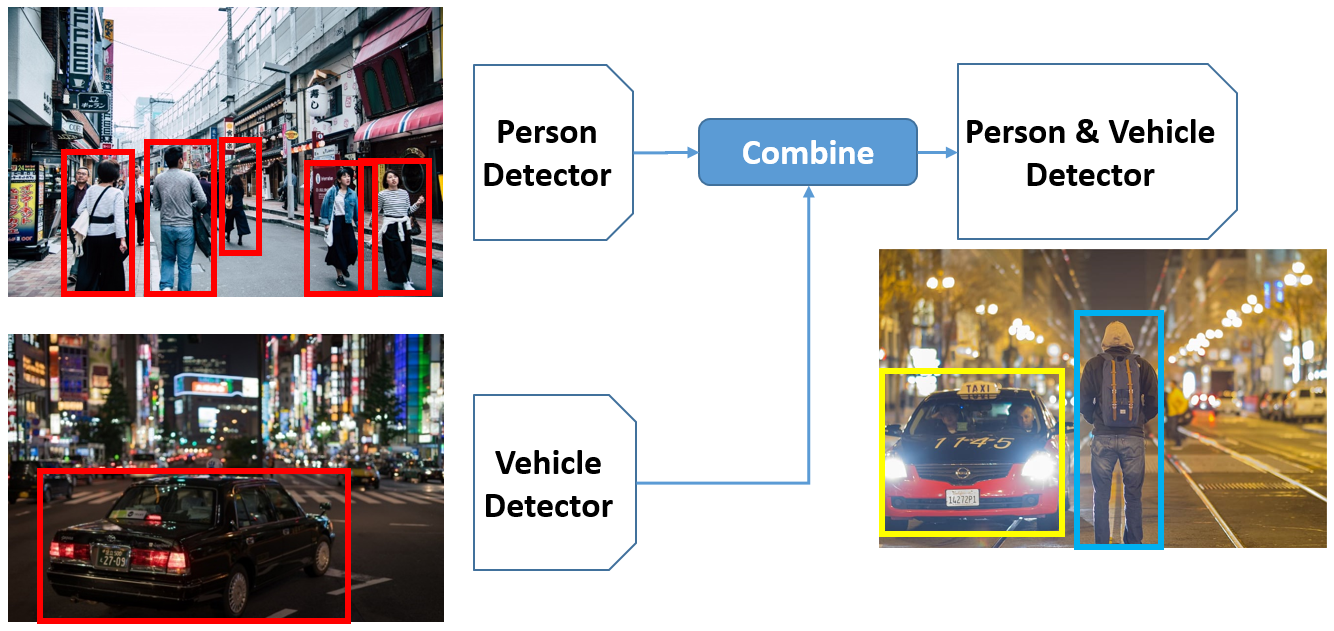}
\vspace{-4pt}
\caption{\small An example of the proposed model composition approach for object detection. Two input models for detecting `person' and `vehicle' categories are combined to create one model that does both.}
\label{fig:example}
\end{SCfigure}

\textbf{Motivation:} A natural question that arises in this case, is whether we can combine the neural networks so that one combined network can perform the same task as several input networks. Fig. \ref{fig:example} shows an example, where two input object detection models to detect `person' and `vehicle' are combined in one model. The benefits of combining models include: a) possible latency improvements due to running one inference as opposed to many, b) in case input models cover partially overlapping or non-overlapping classes/categories, one can build a stronger model with the union of the classes/categories through model composition (i.e. merging models' skills as in Fig. \ref{fig:example}), and c) for applications involving model deployment, e.g. for cloud services providers, it can reduce the deployment frequency/load.

\textbf{Challenges:} Creating a combined model from several input models is a challenging task. First, depending on the target task, the output model may need to have a specific architecture, and not necessarily one that is dictated by the input models. The input models themselves also might have different architectures. Second, in case input models are provided by users of a cloud system, or by different model creators/clients, the individual model owners would likely prefer not to share their training data, labels, not even weights or code. A privacy preserving model composition approach should rely on only a minimum amount of information from the model creators. Third, input models may have only partially overlapping or disjoint class categories. This imposes a major challenge when combining the individual models.

\textbf{Existing methods:} The existing solutions are mostly based on techniques such as knowledge distillation \cite{hinton2015distilling,incrementaldistillation,banitalebidehkordi2021revisiting} or ensembling \cite{zhou2002ensembling}, which may be useful when classes/categories are identical and labeled data are available, but not for the case of arbitrary classes/categories with only unlabeled data. More details regarding the existing approaches are provided in section \ref{section-2}. In summary, to the best of our knowledge, the existing methods do not fully address the three challenges mentioned above.

\textbf{Our contributions:} In this paper, we propose a simple yet effective method to address the model composition of neural networks. Our method supports combination of an arbitrary number of networks with arbitrary architectures. To train a combined model, we leverage the abundance of unlabeled data and having labels or original training data of the input models is not a requirement. However, if any labeled data are available, the algorithm uses them to further boost the performance of the output model. Furthermore, we put no restrictions on the type and number of object categories of the input models. We demonstrate the effectiveness of our method through an extensive set of experiments for the task of object detection.

\vspace{-2pt}
\section{Related works} \label{section-2}
\vspace{-4pt}
Related to our work are the following approaches:\\
\textbf{Network Ensembling:} Ensembling is a common way of aggregating the predictions of more than one models. Ensembling strategies are well explored in the literature \cite{zhou2002ensembling, casadoensemble, solovyev2019weighted}. Simplest ways could be naive averaging of predictions.\\
\textbf{Architectural Combination:} These methods create new architectures from the input models. Adaptive Feeding (AF) \cite{zhou2017adaptive} proposes to use simultaneously two small and large networks that are trained to perform a same task. A linear classifier decides which examples to go to the small or large model. Their goal was to improve the inference speed. In another work, \cite{chou2018unifying} Unifying\&Merging (U\&M) proposes to design a new architecture based on existing input architectures, to support learning multiple tasks.\\
\textbf{MultiTask Networks:} MultiTask networks learn multiple tasks in one model \cite{ruder2017overview,VandenhendeGGB20,JhaKBN20}. Tasks train simultaneously, not by combining already trained individual networks.\\
\textbf{Incremental Learning (IL):} Gradually adding new categories while trying to limit the catastrophic forgetting~\cite{peng2020faster}. \\
\textbf{Dataset Merging:}~\cite{rame2018omnia} Dataset merging is closest work to our study. It proposes to combine datasets by filling the missing annotations of non-overlapping categories. 

While existing works have some correlation to the problem we study, none of them directly addresses this problem. Specifically, our proposed method combines neural networks of same tasks (e.g. classification, detection, etc.) using unlabeled data. If any labels are available, it will use them to further boost the performance. On the other hand, most existing methods mentioned above require labels to be available. In addition, our method supports having overlapping or disjoint target label categories while existing methods such as multi-task networks, U\&M, AF, ensembling, or Dataset Merging only support homogeneous categories. 
Moreover, our method is architecture agnostic. In contrast, AF, MultiTask, IL, and Dataset Merging, don't support arbitrary input model architectures. 

It is also worth noting that most existing methods require access to input model weights or code, to construct a combined model. Methods such as multi-tasking, IL, Dataset Merging or U\&M need full access to input models, in order to design a new combined architecture. Our method only requires an inference API, and thus treats the input models as black boxes, which in turn leads to a better privacy protection for the clients.



\vspace{-6pt}
\section{Model composition strategy} \label{section-3}
\begin{wrapfigure}{r}{0.50\linewidth}
    \centering
    \vspace{-16pt}
    \includegraphics[width=0.45\paperwidth]{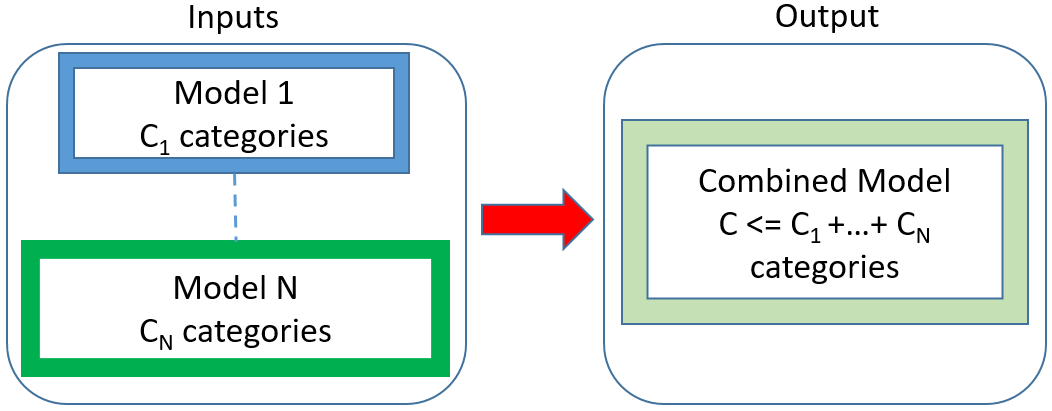}
    \vspace{-18pt}
    \caption{\small Inputs and output of model composition.}
    \label{fig:in-out}
\end{wrapfigure}

\begin{figure}[!b]
\centering\includegraphics[width=0.70\paperwidth]{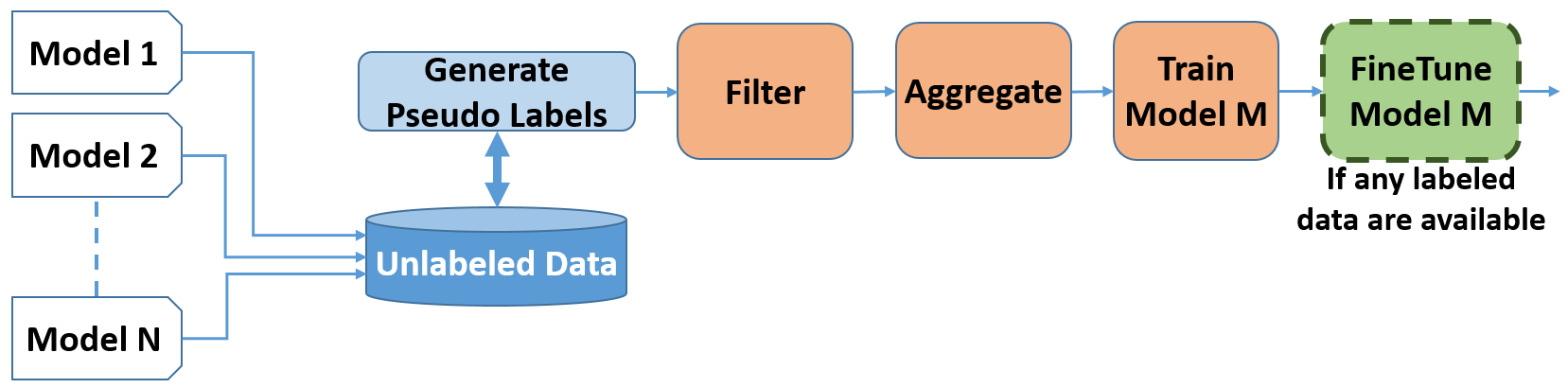}\vspace{-10pt}
\caption{\small Flow-diagram of the model composition method used in our study.}
\label{fig:method}
\end{figure}

\vspace{-8pt}
This section provides details regarding the model composition method we use in this paper. Fig. \ref{fig:in-out} shows the inputs and outputs of this process. In addition, Fig. \ref{fig:method} shows a flow-diagram of different steps within this method. As observed from Fig. \ref{fig:method}, a number of models are provided as inputs. We then collect the predictions of these models over an unlabeled set of images. These predictions are filtered and aggregated to form a set of generated pseudo-labels, which is later used to train the output model $M$. If any labeled data are available, $M$ will be fine-tuned on them. Algorithm \ref{algo:method} shows a break-down of these different steps.

\begin{figure*}
\centering\includegraphics[width=0.75\paperwidth]{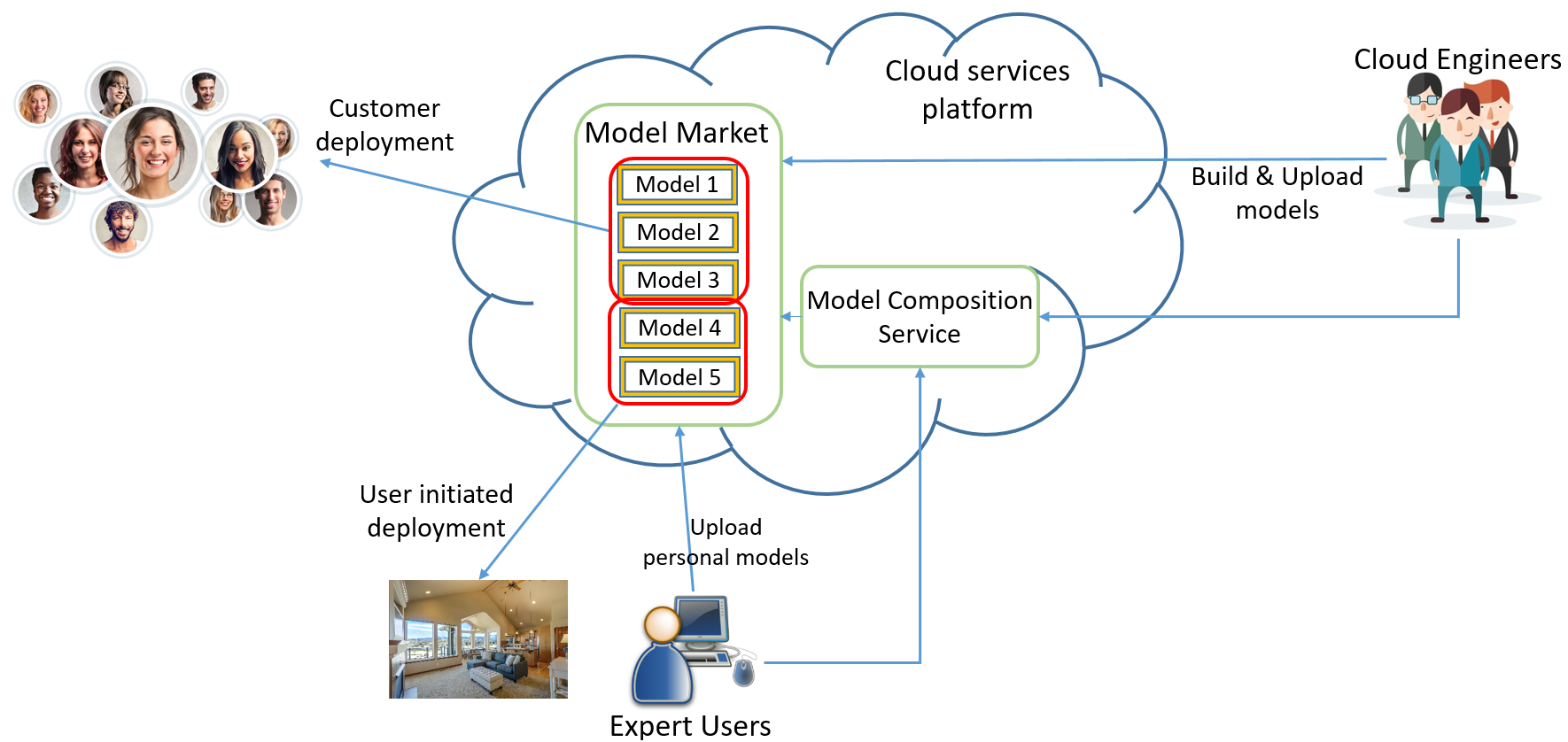}\vspace{-10pt}
\caption{\small An embodiment of how our model composition strategy would be implemented in a cloud services platform. Models shown as groups are combined with each other.}
\label{fig:cloud-embodiment}
\end{figure*}

An embodiment of how our method would be implemented for usage in a cloud services provider platform is demonstrated in Fig. \ref{fig:cloud-embodiment}. As shown in this figure, in the context of a cloud services provider, model composition can be leveraged for: less frequent model/data deployment/transfer, building stronger models, faster overall inference, empowering the model markets, and encouraging users to share their models in an incentive sharing strategy.

\vspace{-4pt}
\subsection{Filtering pseudo-labels}
\vspace{-6pt}
Since the input model predictions are not always perfectly accurate, the generated pseudo-labels will be noisy, and therefore less reliable. These training examples could have an adverse impact on the training of the output model. We filter out such kind of examples, by employing an entropy-based thresholding mechanism.

For a given input $x$ and a network function $f$ such that $p=f(x)$, the entropy is given by $H(f,x)=-\sum_{i} p_{f_i}\log p_{f_i}$. An unreliable pseudo-label may be discarded if $H(f,x) > \tau$, for some threshold $\tau$.
Although entropy thresholding does not guarantee a perfect filtering of bad pseudo-labels, but in practice it works well and has been used as a confidence indicator for similar purposes \cite{teerapittayanon2016branchynet, saporta2020esl, rottmann2018deep}. Note that for some tasks such as object detection, models output a confidence score that can also be used for filtering bad pseudo-labels.

\vspace{-5pt}
\subsection{Aggregation of pseudo-labels}
\vspace{-5pt}
Next, in the pseudo-label aggregation phase, we employ a consensus based strategy, where the majority of the input models need to agree on a pseudo-label in order for it to qualify as a candidate and pass to the next step. Pseudo-label aggregation can be done in various ways such as unanimous (all models agree), affirmative (union of all predictions), consensus (majority voting), etc. \cite{casadoensemble}. Our experiments showed all these methods can be used with minor performance variations. We chose the consensus approach for the experiments since for combining a higher number of models, intuitively it makes more sense (See section \ref{section-4} for a 10-model example). Note that for some tasks such as image classification, the aggregation will be a simple majority voting mechanism. For some other tasks such as object detection, it becomes more complicated due to the nature of the task. Here, we review our method of pseudo-label aggregation for object detection, which can also be extended to other similar tasks such as instance segmentation, tracking, etc.

\textbf{Details of the pseudo-label aggregation strategy:} Let $\cal \overline{D}$ denote the unlabeled dataset used. The input to the pseudo-label aggregation procedure is a list ${\cal S} = 
[\overline{\mathbf{y}}_1, ..., \overline{\mathbf{y}}_N]$, where each $\overline{\mathbf{y}}_i$ itself is a list of detections from an input model over all unlabeled training images in $\cal \overline{D}$. We then create a new list ${\cal S}_{im} = [p_1, ..., p_{|\cal \overline{D}|}]$ so that each $p_i$ contains predictions of all models on one single image, and length of ${\cal S}_{im}$ is equal to number of all images in $\cal \overline{D}$.

\begin{wrapfigure}{r}{0.60\linewidth}
    \centering
    \includegraphics[width=0.60\columnwidth]{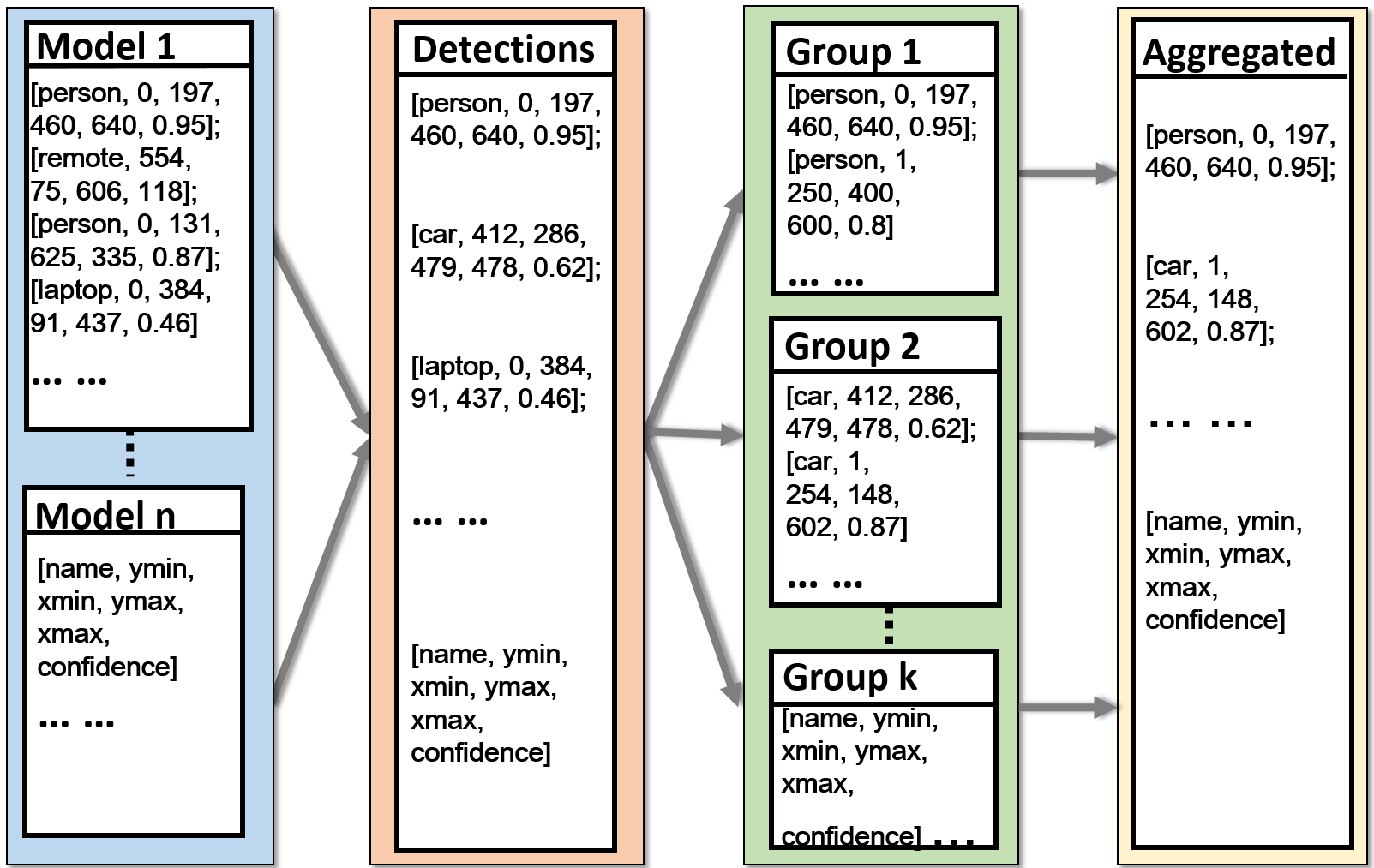}
    \vspace{-20pt}
    \caption{\small Pseudo-label aggregation scheme, an example.}
    \label{fig:ensemble-method-main}
\end{wrapfigure}

Next, for each element $p_i$, we unite the predictions by their category names and the overlapping of their bounding boxes. If the overlapped area of any two elements in $p_i$ is higher than a certain threshold, and meanwhile if these two elements are of the same category, then they are treated as detections of the same object, which are further grouped together into a sub-list: $\overline{p}_{i} = \{\overline{p}_{ij}\}$.
Subsequently, we decide whether to keep each element $\overline{p}_{ij}$ depending on the number of unique models with predictions included in $\overline{p}_{ij}$, denoted by $K_{ij}$. In the most strict case, $\overline{p}_{ij}$ is kept in the list only when  $K_{ij} = N_{ij}$, where $N_{ij} \leq N$ is the maximum number of models that may predict the object category corresponding to $\overline{p}_{ij}$; If we want a majority voting, then $\overline{p}_{ij}$ is kept when $K_{ij} \geq N_{ij}/2$; If a simple stacking strategy is used, then $\overline{p}_{ij}$ is kept regardless of $K_{ij}$. At this point, each $\overline{p}_{ij}$ could still have several candidate detections for the same region. Processing all of them through the detection network is not only cumbersome but could also decrease the overall performance. Therefore, we applied soft non-maximum suppression (Soft-NMS) \cite{bodla2017soft} to each $\overline{p}_{ij}$ to filter the predictions a second time.
Algorithm \ref{algo:aggregation} formally captures these steps and Figure \ref{fig:ensemble-method-main} demonstrates an example.

\textbf{Remark 1}: $\overline{p}_{ij}$ for image $i$, represents a list of bounding boxes predicted on a particular object $j$, i.e. detections of a same object by different models. $K_{ij}$ is the number of unique models in $\overline{p}_{ij}$. $N_{ij}$ is the number of models that have the category of $\overline{p}_{ij}$ in their label set (i.e. number of models that actually have the capability of detecting that object category). As such, in general $K_{ij}\leq N_{ij}\leq N$. In an ideal case where all eligible models can detect an object $ij$, we will have $K_{ij}=N_{ij}$. If all input models have the same category label set, $N_{ij}=N$. In the case input models have different but overlapping categories (i.e. there is at least one category that is not supported by all models), for at least some $i$ and $j$, $N_{ij}<N$. If all models have strictly different categories (no overlap), $K_{ij}=N_{ij}=1$. And finally if some particular categories only belong to one model, for those categories $K_{ij}=N_{ij}=1$.

\begin{figure}
\vspace{-6pt}
{
\scalebox{0.84}{
\begin{minipage}[t]{0.59\linewidth}

\vspace{30pt}
\begin{algorithm}[H]
\caption{\label{algo:method} Our Model Composition Approach}
\begin{algorithmic}

\algnewcommand\algorithmicinput{\textbf{Inputs:}}
\algnewcommand\algorithmicinputopt{\textbf{Optional:}}
\algnewcommand\INPUT{\item[\algorithmicinput]}
\algnewcommand\INPUTopt{\item[\algorithmicinputopt{}]}
\algnewcommand\algorithmicoutput{\textbf{Output:}}
\algnewcommand\OUTPUT{\item[\algorithmicoutput]} 

\INPUT Input models {$\mathbf{M}_{1},..., \mathbf{M}_{N}$}; unlabeled data {$\cal \overline{D}$}

\INPUTopt Output model architecture; labeled data {$\cal D$}

\OUTPUT Combined model {$\mathbf{M}$}
\Procedure{\textbf{M}odel\textbf{C}omposition}{$\mathbf{M}_{1},..., \mathbf{M}_{N}$, {$\cal \overline{D}$}}

\State $\cal S \gets \emptyset$

\For{$i\gets {1,2,...,N}$}
    \State $\{\overline{\mathbf{y}}_i\} \gets GeneratePseudoLabels(\mathbf{M}_i, \cal \overline{D})$
    
    \State $\{\overline{\mathbf{y}}_i\}\ \gets FilterPseudoLabels(\{\overline{\mathbf{y}}_i\})$

    \State {${\cal S} \gets {\cal S} \cup \{\overline{\mathbf{y}}_i\}$}
\EndFor

\State {${\cal \overline{S}} \gets AggregatePseudoLabels({\cal S})$}

\State Initialize $\mathbf{M}$

\For{$j \gets 1,2,...,n_{epochs}$} 

    \For{$B\gets GetBatch(\cal \overline{D}, \cal \overline{S})$}

        \State $\mathbf{M} \gets ApplyGradients(B, \mathbf{M})$
    \EndFor
\EndFor

\EndProcedure
\end{algorithmic}
\end{algorithm}

\end{minipage}\qquad

\begin{minipage}[t]{0.52\linewidth}

\begin{algorithm}[H]
\caption{\label{algo:aggregation} Our Pseudo-Label Aggregation}
\begin{algorithmic}
\algnewcommand\algorithmicinput{\textbf{Inputs:}}
\algnewcommand\INPUT{\item[\algorithmicinput]}
\algnewcommand\algorithmicoutput{\textbf{Output:}}
\algnewcommand\OUTPUT{\item[\algorithmicoutput]} 

\INPUT Pseudo-labels $\cal S$ from $N$ input models;\\ unlabeled data {$\cal \overline{D}$}; aggregation strategy $A$

\OUTPUT Aggregated pseudo-labels list $\overline{\cal S}$
\Procedure{Aggregate}{${\cal S}, N, {\cal \overline{D}}, A$}

    \State ${{\cal S}_{im}} \gets GroupByImages({\cal S}, {\cal \overline{D}})$; \hspace{2pt} ${\overline {\cal S}} \gets \emptyset$
    
    \State Let $\{p_1, p_2, ..., p_{|\cal \overline{D}|}\} \gets {\cal S}_{im}$

    \For{$p_i$ in $\{p_1, p_2, ..., p_{|\cal \overline{D}|}\}$}
        
        \State $\overline{p}_{i_{new}} \gets \emptyset$
        
        \State $\{\overline{p}_{ij}\} \gets GroupByObject(p_i)$
        
        \For{$\overline{p}_{ij}$ in $\{\overline{p}_{ij}\}$}
         
            \State Let $K_{ij} = |UniqueModels \in \overline{p}_{ij}|$
            
            \State $N_{ij}$ = |Models supporting $\overline{p}_{ij}$ class|
        
            \If {$A$ is `unanimous'}
                \State delete $\overline{p}_{ij}$ \textbf{if} $K_{ij} < N_{ij}$
            \ElsIf{$A$ is `consensus'}
                \State delete $\overline{p}_{ij}$ \textbf{if} $K_{ij} < N_{ij}/2$
            \EndIf

            \State $\overline{p}_{ij_{soft}} \gets$ SoftNMS($\overline{p}_{ij}$)
        
            \State $\overline{p}_{i_{new}} \gets \overline{p}_{i_{new}} \cup \overline{p}_{ij_{soft}}$
            
        \EndFor
    
        \State $\overline{\cal S} \gets \overline{\cal S} \cup \overline{p}_{i_{new}}$
    
    \EndFor
\EndProcedure
\end{algorithmic}
\end{algorithm}

\end{minipage}
}
}
\end{figure}


\textbf{Remark 2}: The experiments in Section \ref{section-4} contain various practical scenarios in which different aspects of our method are evaluated. Moreover, Figure \ref{fig:scen-4-classes} in supplementary materials \cite{supplementary} shows a scenario where 10 input models with a diverse count and type of object categories are combined. For example in this figure, `teddy bear' is only in $M_1$, `bicycle' is in $M_1,M_3$, `potted plant' is in $M_2,M_3,M_4$, etc. In addition, we also explore in Section \ref{section-4} the task of combining a face detection model with a mask detection one.

\vspace{-4pt}
\subsection{Training pipeline}
\vspace{-5pt}
Once pseudo-labels are filtered and combined, they will be used to train the output model architecture. Any available labeled data will be used in a final fine-tuning stage to improve the performance. 
Note that pseudo-labels are generated from unlabeled data. This is because input model owners may only share an interface to their model API for inference, not necessarily the weights, code, architecture, training data, or labels, to protect their privacy. We treat the input models as black boxes. In other words, we only pass a set of arbitrary unlabeled images through them, and collect their predictions to use as pseudo-labels. This further allows us to choose an arbitrary architecture and size for the output model that combines the class categories of the input models. Consequently, our model composition method is agnostic to the training hyper-parameters of the input models such as various optimizers, learning rate schedules, batch sizes, etc.

It is also worth noting that this way of creating composite models can help with light domain shifts. As we see in section \ref{section-4}, input models trained on different datasets (for the same task) can still be effectively combined (even with different sets of categories). To what extent exactly our method can robustly support domain shifts remains out of the scope of this work, and we leave that as a future direction.

\vspace{-3pt}
\section{Experiment results and discussions} \label{section-4}


\vspace{-4pt}
\subsection{Experiments setup}
\vspace{-4pt}
\textbf{Selected model architectures}: We have selected the task of object detection as the main experiments task due to its importance and wide-spread usage in practical applications. That being said, we will also provide results on the task of image classification, as it is often used as a baseline experiment task. For object detection, we utilized the following architectures: EfficientDet-D0~\cite{tan2020efficientdet}, EfficientDet-D1~\cite{tan2020efficientdet}, and RetinaNet-ResNet-50~\cite{lin2017focal}. For the classification task, we used: ResNet-18~\cite{resnet}, ResNet-152\cite{resnet}, and DenseNet-121\cite{densenet}. 

\textbf{Datasets}: We used three sets of benchmarking datasets for object detection: COCO~\cite{COCO}, Pascal-VOC~\cite{Pascal-VOC}, and Open-Images-V5~\cite{OID} (referred by OID hereafter). For classification, we use Caltech-256~\cite{Caltech-256} and OID datasets.

\textbf{Evaluation metrics}: We follow the common practice by using the mean Average Precision, mAP @IoU=0.50:0.95, as the main metric to evaluate the performance of object detection models. We report top-1 accuracy for classification.

\textbf{Training protocols and settings:}
We adopt code from \cite{EffDet-repo} for the object detection experiments, and use the same training hyper-parameters with ImageNet \cite{imagenet} pre-trained backbones. We trained all the models using SGD with a momentum of 0.9. We increased the learning rate from $8e^{-3}$ to $8e^{-2}$ for the first epoch and then trained the remaining 300 epochs using a cosine decay rule with a moving average decay set at 0.9998. Soft NMS was utilized to filter the pseudo-label detections in our method. We used an IoU threshold of 0.5 and a confidence threshold of 0.001. 
For the classification experiments, models were trained for 200 epochs, using an in-house code-base. An SGD optimizer with momentum 0.9 was used, and learning rate was exponentially increased from 0 to 0.01 for the first 8 epochs and then annealed down exponentially to 0.0001 in the remaining epochs.

\vspace{-2pt}
\subsection{Object detection results} \label{sec:obj-results}
\vspace{-4pt}
Our experiments are categorized in various scenarios, which are explained in this subsection. These scenarios cover various possible cases of input models' architectures, training data, and what kind of unlabeled data were used in our algorithm. Table \ref{tab:datasplits-and-models} provides a summary of these scenarios. In Table \ref{tab:datasplits-and-models}, training data in each case is constructed from the training set of VOC, COCO, OID, or a subset of them (unlabeled). Validation sets are also built from the validation sets of VOC and COCO: a subset of COCO (union of input categories) for scenario 1, union of the val set of COCO \& VOC for scenario 2 \& 3, and val set of COCO for scenario 4. As such, validation sets may be different across different scenarios, but are the same within one scenario. Moreover, the class distributions of data for scenario 1 \& 4 are shown in Table \ref{tab:scen-1-classes} and Figure \ref{fig:scen-4-classes} (supplementary), and for scenario 2 \& 3 it follows the distributions of COCO \& VOC. Next, we will go over the details of each experiment.

\textbf{Scenario 1: Combining detectors with different expertise}: We took 3 models, each trained on a subset of the COCO dataset but designed for a different purpose, one for detection of transportation related objects, one for sports related, and the other for home objects. These categories have some partial overlap. Table \ref{tab:scen-1-classes} shows the object categories used for each model. The combined model achieved by our model composition procedure combines the skills of the input models, and builds a stronger model with all object categories. We tried our method in two ways: one using unlabeled COCO images (similar data distribution to training data, but without using the labels), and the other using unlabeled OID (open images) dataset (entirely different dataset with different distribution). The upper-bound of the performance would be to train a model with all labels of all object categories (supervised). This model achieved 35.11\% mAP on validation set of COCO (considering only object categories corresponding to the ones it was trained on). On the same validation set, our method achieved 32.61\% when using unlabeled COCO, and 30.97\% when using unlabeled OID. This shows that our method can effectively combine the models with different expertise, and achieve a performance close to that of the supervised upper-bound model. We further investigated the performance of our method if partial labels are available for fine-tuning, in a semi-supervised manner. Table \ref{tab:OD-results} shows the results for this experiment. As observed, with fine-tuning, our method could even surpass the supervised model with 100\% of labels. 

\begin{table}
\centering
\fontsize{7.5}{8}\selectfont
\begin{tabular}[t]{p{1.4cm}p{2cm}p{2.5cm}p{2.7cm}p{1.3cm}}
\toprule
Experiment & Architecture & Model/Method & Train Set & Size\\
\midrule
& \hspace{1em}EffDet-D0 & input (supervised) & COCO subset 1 & 72K \\
& \hspace{1em}EffDet-D0 & input (supervised) & COCO subset 2 & 66K \\
& \hspace{1em}EffDet-D0 & input (supervised) & COCO subset 3 & 81K \\
\textbf{Scenario 1} & \hspace{1em}EffDet-D0 & Upper-bound & COCO subsets union & 89K \\
& \hspace{1em}EffDet-D0 & ModelComp (Ours) & unlabeled COCO & 118K\\
& \hspace{1em}EffDet-D0 & ModelComp (Ours) & unlabeled OID$^*$ & 1.9M\\\cmidrule{2-5}


& \hspace{1em}EffDet-D0 & input (supervised) & COCO & 118K \\
& \hspace{1em}EffDet-D0 & input (supervised) & VOC & 17K \\
\textbf{Scenario 2} & \hspace{1em}EffDet-D0 & Upper-bound & COCO+VOC & 135K \\
& \hspace{1em}EffDet-D0 & ModelComp (Ours) & unlabeled COCO+VOC & 135K\\
& \hspace{1em}EffDet-D0 & ModelComp (Ours) & unlabeled OID$^*$ & 1.9M\\\cmidrule{2-5}

& \hspace{1em}EffDet-D1 & input (supervised) & COCO & 118K \\
& \hspace{1em}EffDet-D0 & input (supervised) & VOC & 17K \\
\textbf{Scenario 3} & \hspace{1em}RetinaNet-R50 & Upper-bound & COCO+VOC & 135K \\
& \hspace{1em}RetinaNet-R50 & ModelComp (Ours) & unlabeled COCO+VOC & 135K\\
& \hspace{1em}RetinaNet-R50 & ModelComp (Ours) & unlabeled OID$^*$ & 1.9M\\\cmidrule{2-5}

& \hspace{1em}EffDet-D0 & 10 inputs (supervised) & 10 COCO partitions & $\approx$12K each\\
& \hspace{1em}EffDet-D0 & Upper-bound & COCO & 118K \\
\textbf{Scenario 4} & \hspace{1em}EffDet-D0 & ModelComp (Ours) & unlabeled COCO & 118K\\
& \hspace{1em}EffDet-D0 & ModelComp (Ours) & unlabeled OID & 100K\\
\bottomrule
\end{tabular}
\caption{\label{tab:datasplits-and-models}Data-splits and models for object detection. 
{\small *We also evaluate on a 118K subset of OID.}}
\end{table}


\begin{table}
\centering
\fontsize{7.4}{8}\selectfont
\begin{tabular}[t]{p{1.2cm}p{10.8cm}}
\toprule
Model skill &  Categories supported\\
\midrule
Transportation & person, bicycle, car, motorcycle, bus, truck, traffic light, fire hydrant, stop sign, parking meter\\
Sports & person, bicycle, frisbee, skis, snowboard, sports ball, skateboard, baseball bat, baseball glove, motorcycle\\
Home & person, bicycle, chair, couch, bed, dining table, skateboard, refrigerator, toilet, tv\\
\bottomrule
\end{tabular}
\caption{\label{tab:scen-1-classes}\small Object categories for input models of scenario 1.}
\end{table}

\begin{table}
\centering
\fontsize{8}{8}\selectfont
\begin{tabular}[t]{lllllllll}
\toprule
& \multicolumn{1}{c}{ } & \multicolumn{7}{c}{Proportion of labeled data used (\%)} \\
\cmidrule(l{3pt}r{3pt}){3-9}
Experiment & \hspace{3em}Method & 0 & 1 & 5 & 10 & 30 & 50 & 100 \\
\midrule

& \hspace{0.em}Supervised: COCO$_{subsets}$ & - & 3.9 & 16.4 & 20.3 & 28 & 30.8 & 35.1 \\
\textbf{Scenario 1} & \hspace{0.em}Ours: COCO$^U$+FT & \textbf{32.6} & 32.6 & 33.5 & 33.7 & 34 & 34 & 35.7 \\
& \hspace{0.em}Ours: OID$^{U-118k}$+FT & 28.5 & 29.1 & 29.6 & 31.8 & 32.8 & 33.9 & 35.3 \\
& \hspace{0.em}Ours: OID$^U$+FT & 31 & 31 & 31.3 & 32.3 & 33.9 & 34.6 & \textbf{36} \\\cmidrule{2-9}

& \hspace{0.em}Supervised: COCO+VOC & - & 5.3 & 16.3 & 19.7 & 25 & 27.5 & 32.9\\
\textbf{Scenario 2} & \hspace{0.em}Ours: [COCO+VOC]$^U$+FT & \textbf{29} & 29.2 & 29.3 & 30 & 30.3 & 30.4 & \textbf{33.1}\\
& \hspace{0.em}Ours: OID$^{U-118k}$+FT & 26 & 26.5 & 27 & 27.9 & 29.2 & 30.3 & 32.5\\
& \hspace{0.em}Ours: OID$^U$+FT & 27.4 & 27.4 & 27.8 & 29 & 30 & 30.5 & 32.7\\\cmidrule{2-9}

& \hspace{0.em}Supervised: COCO+VOC & - & 4 & 12.6 & 18.5 & 27.5 & 30.6 & 35\\
\textbf{Scenario 3} & \hspace{0.em}Ours: [COCO+VOC]$^U$+FT & \textbf{34} & 34.2 & 34.6 & 35 & 35.4 & 35.9 & \textbf{38}\\
& \hspace{0.em}Ours: OID$^{U-118k}$+FT & 16 & 22.5 & 24.9 & 26 & 28.1 & 30.2 & 33.9\\
& \hspace{0.em}Ours: OID$^U$+FT & 16 & 22.6 & 25.2 & 26.6 & 29 & 30.4 & 34.1\\\cmidrule{2-9}

& \hspace{0.em}Supervised: COCO & - & 1.2 & 15.6 & 19.2 & 24.4 & 27.9 & \textbf{33.6}\\
\textbf{Scenario 4} & \hspace{0.em}Ours: COCO$^U$+FT & \textbf{24.4} & 24.5 & 26.7 & 27.7 & 28.6 & 29.1 & 33.1\\
& \hspace{0.em}Ours: OID$^U$+FT & 16.6 & 16.8 & 19.9 & 21.6 & 25.2 & 27.1 & 32.4\\
\bottomrule
\end{tabular}
\caption{\label{tab:OD-results}\small Object detection results: FT, Ours, and $U$, refer to fine-tuning, Model Composition, and unlabeled, respectively. Combined models, even without any labels, show a competitive performance.}
\end{table}

\textbf{Scenario 2: Combining input models that are trained on entirely different datasets.} In scenario 1, input models had different expertise, by getting trained on different subsets of COCO (examples roughly came from a similar distribution). Scenario 2 investigates a more challenging case, where input models were trained with data from entirely different datasets, hence different distributions. To this end, we trained input models on COCO and Pascal-VOC datasets respectively. Similar to the previous scenario, we studied two choices of unlabeled data for our Model Composition method: a) unlabeled data from the same distribution as training data (in this case COCO+VOC images without using labels), and b) unlabeled data from a different dataset all together, e.g. the OID dataset. Note that the input models were trained on a different number of object categories (with overlap), and the output combined model was trained to support the union of object categories of the input models.

Table \ref{tab:OD-results} shows the results of this experiment. It is observed from Table  \ref{tab:OD-results} that in the unsupervised case (i.e. no labeled data was used), our method achievs 29\% and 27.4\% mAP, close to the fully supervised performance of the upper-bound model. We also see from Table  \ref{tab:OD-results} that when partially labeled data are used for further fine-tuning, our method shows significant improvements over supervised training. In particular, when using 1\%, 5\%, and 10\% of labels, our method shows +22.1\%, +13\%, and +10.3\% gaps over supervised training.



\textbf{Scenario 3: Combining input models with different architectures, that are trained on entirely different datasets.}
In this scenario, we studied the most generic case, in which input models have different architectures, are trained on different datasets, and with a different number of object categories. The output model also was chosen to have a different architecture than the input models (See Table \ref{tab:datasplits-and-models}). This scenario evaluated whether our method can combine the knowledge of models trained on different circumstances, data, and architecture, to a desired new and different architecture. 

Table \ref{tab:OD-results} shows the results of this experiment. It is observed from Table \ref{tab:OD-results} that our method is very effective, and in some cases performs even better than supervised training with 100\% of labels. When partial labels are available for fine-tuning, our method shows a strong performance, with large gaps compared to supervised training, especially in the low label range. Moreover, Table \ref{tab:OD-results} shows that unsupervised training with our method achieved an mAP of 34\%, only 1\% below supervised training with all labels. After fine-tuning, we were able to meet the same performance as supervised training with only 10\% of the original data.



\textbf{Scenario 4: Having a large number of input models.} This scenario investigates the case when a larger number of input models are provided. This would increase the diversity among the models since they can be trained on different data, or object categories, and thus results in a more challenging situation. To this end, we assumed 10 input models. Each model was trained on a randomly selected subset of the COCO dataset, so that training data for each model had no overlap to the other models. However, object categories could have overlap, as their type and count were chosen randomly. Supplementary materials \cite{supplementary} provides a visualization of the type and count of the object categories used for these 10 models. Since each model was trained with roughly 10\% of the COCO training set, different number of object categories for different models resulted in a different per-class size of training data. The imbalance here made model composition harder, but mimicked realistic situations where training data can in fact be imbalanced for input models. As mentioned, for these 10 models, categories were randomly selected and the number of categories was selected from 5, 10, 20, 30, and 40. Note that generating 10$\times$ pseudo-labels on unlabeled data can be time-consuming (although it can be parallelized in production). Therefore, we only used 100K randomly selected examples from the OID dataset for this experiment.



We observe from Table \ref{tab:OD-results} that our model composition method can effectively combine the 10 input models into a single new model with the union of their object categories.

\begin{table}[!b]
\centering
\fontsize{8}{8}\selectfont
\begin{tabular}[t]{cccp{0.8cm}}
\toprule
Model/Expertise &  Train set & Validation set & AP(\%)\\
\midrule
input: Face (D0) & face data 1 (20007) & face data 1 (4079) & 52.29\\
input: Face (D0) & MAFA-faces (30870) & MAFA-faces (5338) & 44.86\\
input: Mask (D1) & MAFA-masks (30870) & MAFA-masks (5338) & 29.63\\ \midrule
ModelComp (R50): w/o filtering \& aggregation & face+mask (50877) & face+mask (9417) & 30.72\\
ModelComp (R50): w/o aggregation & face+mask (50877) & face+mask (9417) & 34.48\\ \midrule
Ours, ModelComp: Face \& Mask (R50) & face+mask (50877) & face+mask (9417) & \textbf{38.90}\\
\bottomrule
\end{tabular}
\caption{\label{tab:face-mask}\small Combining face and mask detectors with different architectures trained on different datasets. Face data 1: Face images from WIDERFACE \cite{wider} and medical masks datasets \cite{med-masks-1,med-masks-2} (including both faces with-masks and without-masks). MAFA faces and masks are obtained from the MAFA dataset \cite{mafa}. We also provide an ablation on the filtering and aggregation modules.}
\end{table}

\textbf{Remark 3: A note on the unsupervised performance of OID:}
As observed in Table \ref{tab:OD-results}, in the challenging scenarios of 3 and 4, the unsupervised (0\% labels) performance of model composition with OID$^U$ is considerably lower that that of COCO$^U$ or [COCO+VOC]$^U$. In this regard, there are a few points worth a mention:
\vspace{-7pt}
\begin{itemize}[leftmargin=*]
    \item In general, using unrelated arbitrary data is expected to result in a lower performance compared to using data from the same distribution as the input models' train set, since pseudo-labels will be less reliable. This is exacerbated in challenging tasks such as scenario 3 where input models are trained on different data and have different architectures with respect to each other and the output model; or in scenario 4 where there are a large number of input models trained on different small-scale data.
    \vspace{-7pt}
    \item It is worth reminding that the case of purely unsupervised model composition means combining an arbitrary number of black-box models (trained on arbitrary data with arbitrary architecture or categories), all without using any labels. In that sense, the real baseline to compare against is the supervised training, which performs much worse than model composition in low data regimes, even in the case of unrelated OID$^U$ data. 
    \vspace{-8pt}
    \item Moreover, the main goal of the paper is to explore whether or not neural networks can be combined using only unlabeled data, and if yes, to what extent (hence the title). We observe from the results that the answer is for the most part yes; however, in case unlabeled data from the original distributions was not available, for some challenging scenarios a small percentage of labels may be needed to achieve a decent performance. 
    \vspace{-8pt}
    \item In a completely unsupervised setting, model composition can still effectively combine input models. The performance will be improved if the unlabeled set size is larger.
\end{itemize}
\vspace{-6pt}

\textbf{Remark 4: A note on practical applications:}
As mentioned in Section \ref{section-3}, a fundamental motivation for our work is a cloud services application, as shown in Figure \ref{fig:cloud-embodiment}, in which engineers and expert users can leverage a model composition service to build stronger models with combined skills, especially in the presence of a large variety of trained models and datasets on the cloud. Different scenarios in the experiments were also inspired by such a philosophy, but designed at various levels of difficulty. 
Here, we add a new practical use-case. In this new experiment, we combine separate models of face and mask detection, to build one that is suitable for both face \& mask detection. Results are shown in Table \ref{tab:face-mask}.

\begin{wraptable}{r}{5.55cm}
\centering
\vspace{-1pt}
\fontsize{8}{8}\selectfont
\begin{tabular}[t]{p{1.2cm}p{0.3cm}p{0.3cm}p{0.3cm}p{0.3cm}p{0.3cm}}
\toprule
\#Categoreis &  5 & 6 & 7 & 10 & 15\\
\midrule
ModelComp & 27.1 & 26.8 & 26.6 & 24.7 & 23.9\\
\bottomrule
\end{tabular}
\vspace{2pt}
\caption{\label{tab:class-numbers}\small Ablation on classes. Details in \ref{sec:obj-results}.}
\end{wraptable}

\textbf{Ablation on the number of categories:}
Next, we study the impact of the number of class categories in the performance. To this end, we take two input models from scenario 4, and combine them with a varying number of classes. $M_1$ is trained on 5 and $M_2$ is trained on 10 object categories (see Figure \ref{fig:scen-4-classes} in supplementary). Each time we add a number of random categories of $M_2$ to $M_1$, so the combined model can have 5,6,...,15 classes. Table \ref{tab:class-numbers} shows the results. Note that in each case the validation/training set will be different as it includes images of a particular set of categories. $M_1$ and $M_2$ have mAP of 27.1\% \& 25.4\%, respectively (each has roughly 12K training, and 1K validation examples). In general, higher number of classes results in slightly lower mAP, but we should also note that unlabeled set size becomes also larger (i.e. more pseudo-labels).

\vspace{-4pt}
\subsection{Image classification results}
\vspace{-4pt}
In addition to our main results on the task of object detection, we also provide a highlight of our results on the task of image classification. Similar to object detection, we designed the classification experiments in the form of different scenarios. 

\noindent\textbf{Scenario 1:} 3 input models, ResNet-18, each trained on 1/3 of the Caltech-256 dataset. 

\noindent\textbf{Scenario 2:} 3 models, ResNet-18, ResNet-152, and DenseNet-121, trained on Caltech-256.

In both scenarios, we tried model composition with unlabeled data from the Caltech-256 dataset (i.e. similar data distribution but without labels), and a 160K subset unlabeled data from OID (i.e. a different dataset altogether). Table \ref{tab:classification-results} shows the results for these scenarios.

In Table \ref{tab:classification-baselines}, we provide a comparison between our method and two additional baselines: i) a simple model ensemble by aggregating directly the prediction of the input models; 
ii) knowledge distillation when using the input models as teachers, such as \cite{hinton2015distilling,ahn2019variational}. For the second baseline, we consider the vanilla distillation \cite{hinton2015distilling} but with soft labels.

It is observed from the results that the proposed method is effective in combining image classification models. In both the unsupervised and semi-supervised cases, our method performs competitively compared to supervised models, even when 100\% of labels are used.

\begin{table}[h]
\centering
\fontsize{8}{8}\selectfont
\begin{tabular}[t]{p{1.4cm}p{2.65cm}p{0.7cm}p{0.7cm}p{0.7cm}p{0.7cm}p{0.7cm}p{0.7cm}}
\toprule
& \multicolumn{1}{c}{ } & \multicolumn{6}{c}{Proportion of labeled data used (\%)} \\
\cmidrule(l{3pt}r{3pt}){3-8}
Experiment & \hspace{2em} Method & 0 & 1 & 5 & 10 & 50 & 100 \\
\midrule
& \hspace{0em}Supervised: Caltech & - & 16.2 & 44 & 61 & 79.4 & 82.4\\
\textbf{Scenario 1} & \hspace{0em}Ours: Caltech$^U$+FT & \textbf{83} & 82.6 & 82.8 & 82.9 & 83 & \textbf{83.5}\\
& \hspace{0em}Ours: OID$^U$+FT & 69 & 70 & 71.2 & 72.9 & 79 & 81.6\\\cmidrule{2-8}
& \hspace{0em}Supervised: Caltech & - & 16.2 & 43.9 & 60.9 & 79.4 & 82.4\\
\textbf{Scenario 2} & \hspace{0em}Ours:  Caltech$^U$+FT & \textbf{83.2} & 82.1 & 81.8 & 81.8 & 83.1 & \textbf{83.3}\\
& \hspace{0em}Ours: OID$^U$+FT & 71.6 & 68.5 & 71.5 & 72.4 & 78.8 & 81.6\\
\bottomrule
\end{tabular}
\caption{\label{tab:classification-results}\small Image classification results: FT, Ours, and $U$, refer to fine-tuning, Model Composition, and unlabeled, respectively. Combined models, even without any labels, show a competitive performance.}
\end{table}

\begin{wraptable}{r}{5.8cm}
\centering
\vspace{-36pt}
\fontsize{8}{8}\selectfont
\begin{tabular}[t]{p{2.06cm}p{1.2cm}p{1.2cm}}
\toprule
 &  Scenario 1 & Scenario 2\\
\midrule
Ensembling & 53.7 & 59.8\\
Vanilla distillation & 64.2 & 68.2 \\
Ours: OID$^U$ & 69 & 71.6\\
\bottomrule
\end{tabular}
\vspace{2pt}
\caption{\label{tab:classification-baselines}\small More classification results on OID$^U$.}
\end{wraptable}

\vspace{-6pt}
\section{Conclusion} \label{section-5}
\vspace{-6pt}
This paper proposed a method for combining multiple trained neural networks into a single model, using unlabeled data. To this end, first the input models’ predictions (pseudo-labels) were collected. The pseudo-labels were then filtered based on confidence scores of the predictions. Next, a consensus aggregation strategy was incorporated to combine these pseudo-labels. The remaining pseudo-labels were used to train the output model. The proposed method supported the use of an arbitrary number of input models with arbitrary architectures and categories. Performance evaluations on various datasets, tasks, and network architectures demonstrated the effectiveness of the proposed method.

\balance

\end{spacing}

\clearpage
\begin{spacing}{0.94}
{\small
\bibliography{references_short}
}
\end{spacing}

\clearpage

\section{Supplementary materials}
This section contains the supplementary materials.

\subsection{Source code}
We share our implementation code to make it easy to reproduce our results. The source-code is attached to the supplementary materials in a `code' directory. We also provide detailed instructions for training and evaluating our models in `README.md' files.

\subsection{Additional visualizations}
\begin{figure}[!b]
\centering\includegraphics[width=0.95\columnwidth]{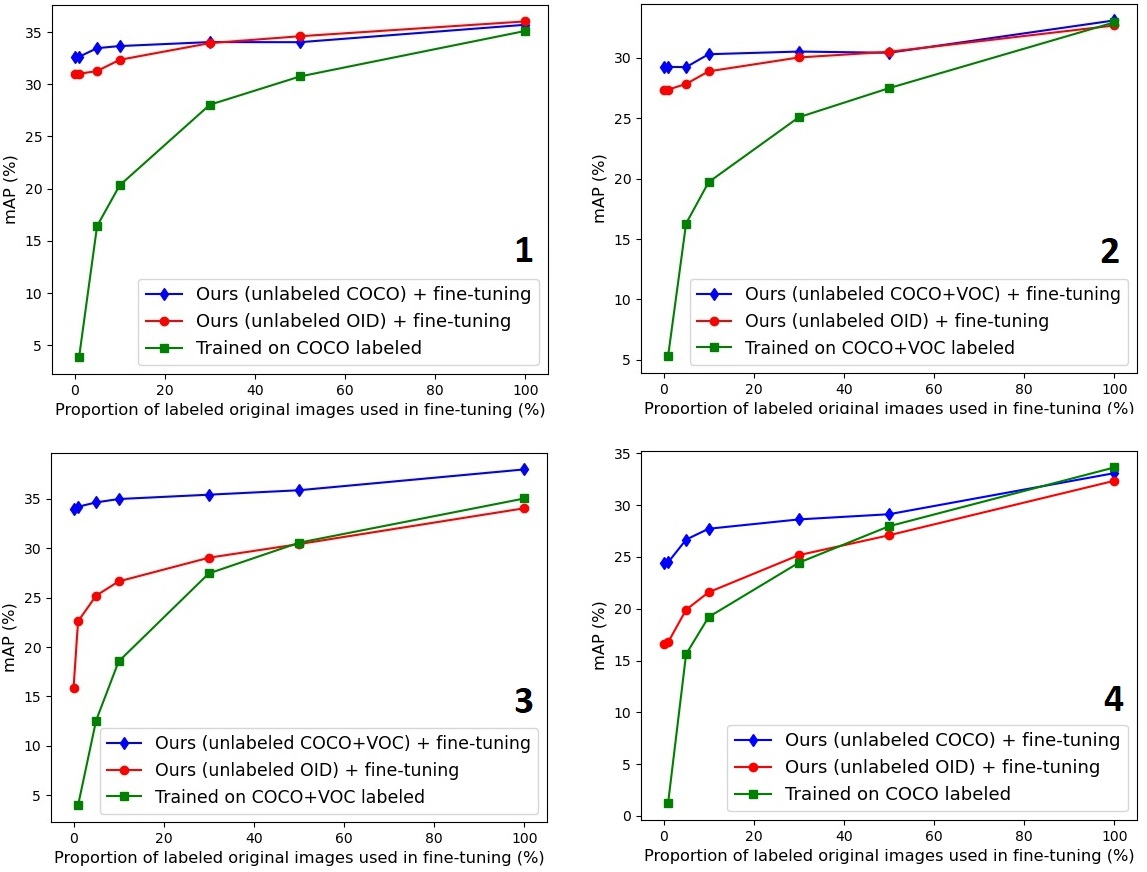}
\vspace{-10pt}
\caption{\small Results of different scenarios for fine-tuning. Each scenario is tagged with its number.}
\label{fig:result-4}
\end{figure}

\begin{figure*}[!b]
\centering\includegraphics[width=0.8\paperwidth]{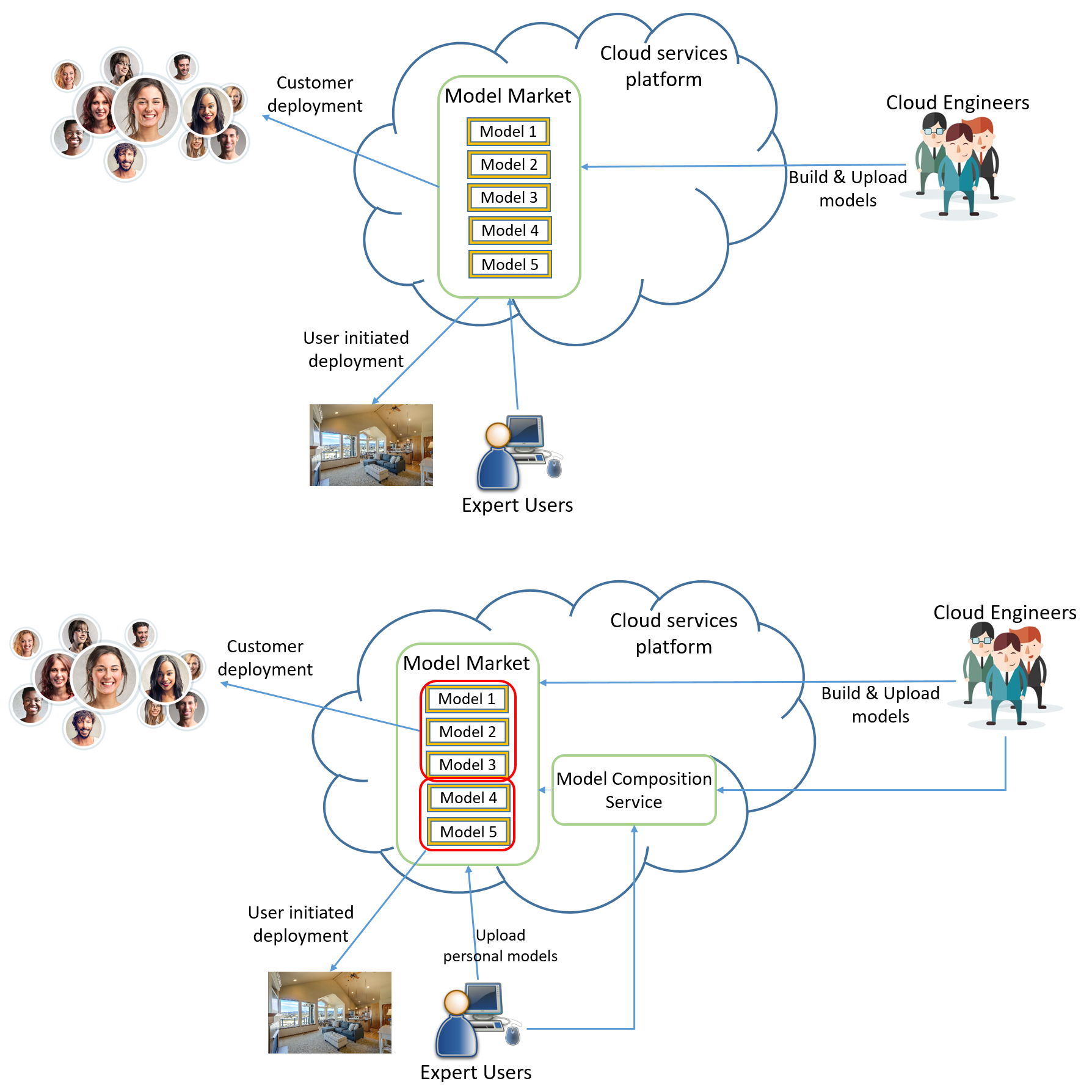}\vspace{-10pt}
\caption{\small An embodiment of how our model composition strategy would be implemented in a cloud services platform. Top: existing system. Bottom: after applying model composition.}
\label{fig:cloud-embodiment-full}
\end{figure*}

\begin{figure}
\centering\includegraphics[width=0.9\columnwidth]{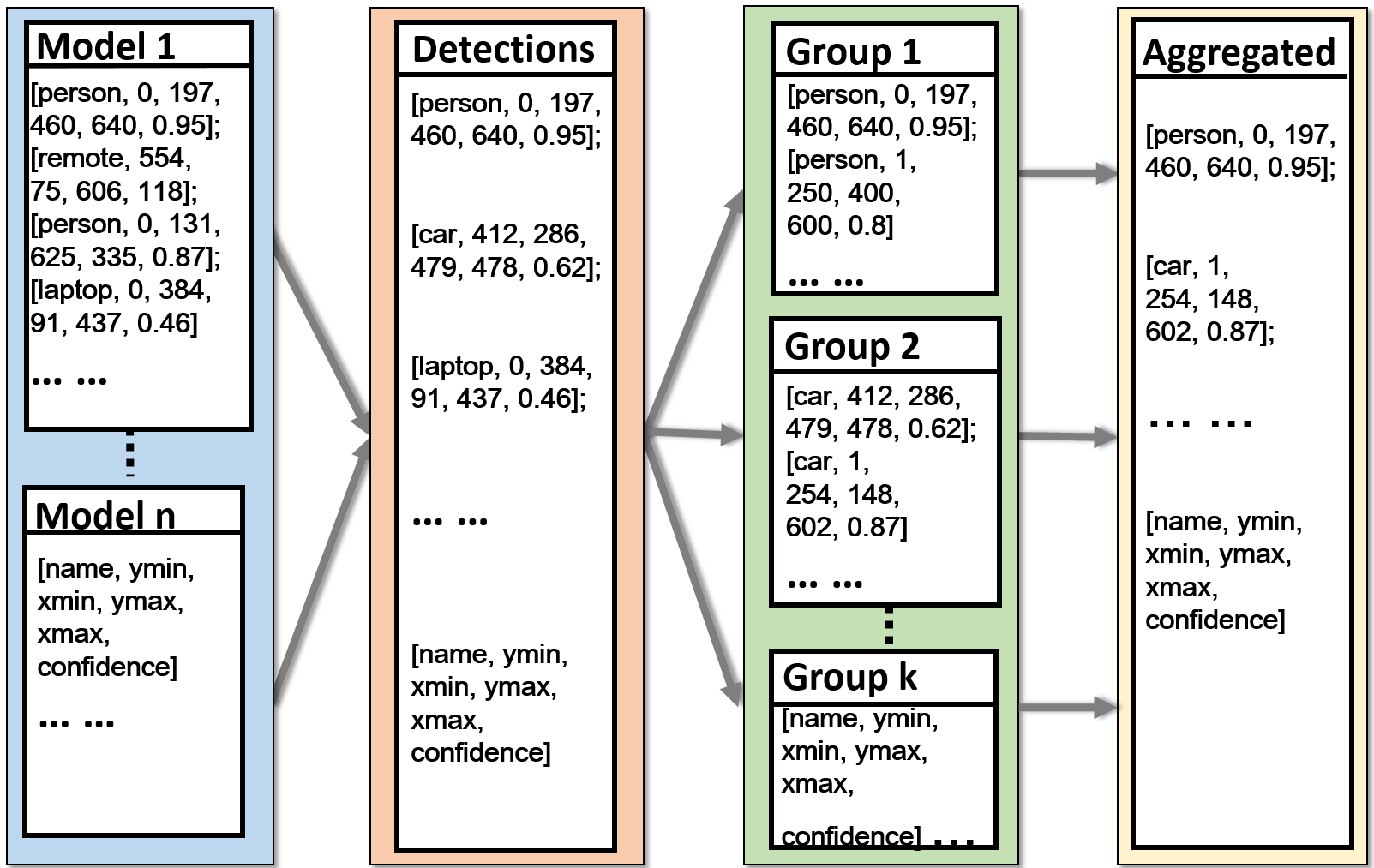}
\vspace{-8pt}
\caption{\small Pseudo-label aggregation flow diagram, an example.}
\label{fig:ensemble-method}
\end{figure}

\begin{figure}
\centering\includegraphics[width=0.98\columnwidth]{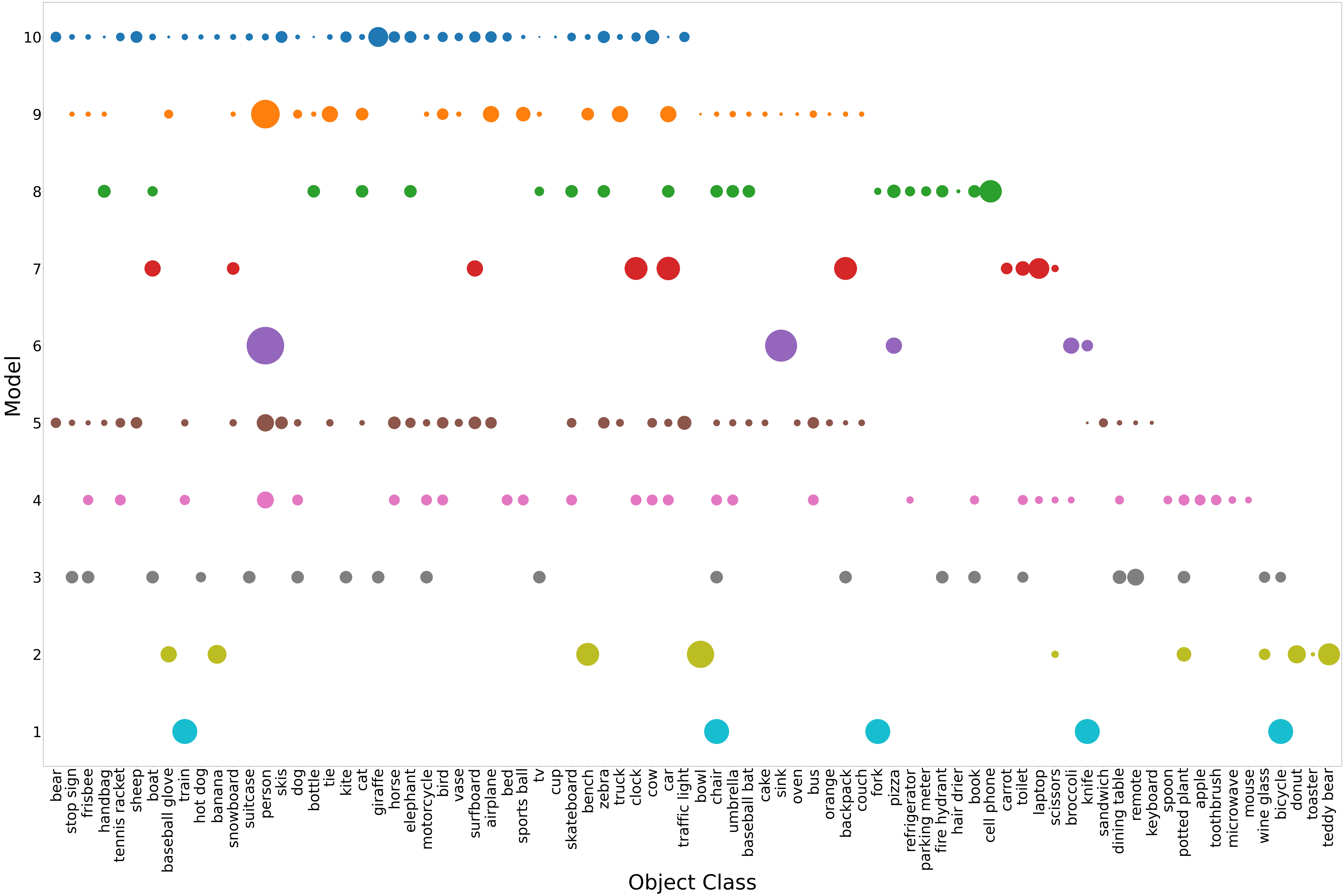}
\vspace{-8pt}
\caption{\small Object categories and data size for models used in object detection scenario 4. Larger dots correspond to higher number of examples. 10 rows correspond to 10 different models.}
\label{fig:scen-4-classes}
\end{figure}

Fig. \ref{fig:result-4} provides a visualization of object detection results of Table \ref{tab:OD-results}. We observe from this figure that in low data regimes model composition performs considerably better than supervised training with partial data. Fig. \ref{fig:cloud-embodiment-full} shows an extended visualization on the cloud embodiment introduced in the paper. In this figure, we provide an easier comparison between before \& after incorporating the model composition as a service. 
Moreover, Fig. \ref{fig:ensemble-method} demonstrates an example of pseudo-label aggregation procedure of Algorithm \ref{algo:aggregation}. 
In addition, Fig. \ref{fig:scen-4-classes} visualizes the data splits of object detection scenario 4, where we combined 10 models trained on different COCO subsets.

\end{document}